\documentclass[final,5p,times,twocolumn]{elsarticle}

\usepackage{amssymb}
\usepackage{amsmath}
\usepackage{booktabs}
\usepackage{amsthm}
\theoremstyle{plain}
\usepackage{graphics}
\usepackage{epstopdf}
\usepackage{epsfig}
\usepackage{lineno}
\usepackage{array}
\usepackage{amsmath,amssymb}
\usepackage{mathrsfs}
\usepackage{graphicx}
\usepackage{caption}
\usepackage{multirow}
\usepackage{subcaption}
\usepackage{bm}
\usepackage{xcolor,colortbl}
\definecolor{iccvblue}{rgb}{0.21,0.49,0.74}
\usepackage[pagebackref,breaklinks,colorlinks,allcolors=iccvblue]{hyperref}

\begin{document}

\begin{frontmatter}



\title{AnchorFormer: Differentiable Anchor Attention for Efficient Vision Transformer}



\author[PetroChina]{Jiquan Shan}
\author[PetroChina]{Junxiao Wang}
\author[PetroChina]{Lifeng Zhao}
\author[PetroChina]{Liang Cai}
\author[HKU]{Hongyuan Zhang}
\author[EASA,SCUT]{Ioannis Liritzis \corref{mycorrespondingauthor}}

\address[PetroChina]{PetroChina Changqing Oilfield Company, Xi'an, Shaanxi, China}
\address[HKU]{The University of Hong Kong, Hong Kong}
\address[EASA]{European Academy of Sciences and Arts, Austria}
\address[SCUT]{South China University of Technology, Guangzhou, Guangdong, China}


\cortext[mycorrespondingauthor]{Corresponding author}

\begin{abstract}
    Recently, vision transformers (ViTs) have achieved excellent performance on vision tasks by measuring the global self-attention among the image patches. 
    Given $n$ patches, they will have quadratic complexity such as $\mathcal{O}(n^2)$ and the time cost is high when splitting the input image with a small granularity. 
    Meanwhile, the pivotal information is often randomly gathered in a few regions of an input image, some tokens may not be helpful for the downstream tasks. 
    To handle this problem, we introduce an anchor-based efficient vision transformer (\textbf{AnchorFormer}), which employs the anchor tokens to learn the pivotal information and accelerate the inference. Firstly, by estimating the bipartite attention between the anchors and tokens, the complexity will be reduced from $\mathcal{O}(n^2)$ to $\mathcal{O}(mn)$, where $m$ is an anchor number and $m < n$. 
    Notably, by representing the anchors with the neurons in a neural layer, \textcolor{black}{we can differentiably learn these anchors} and approximate global self-attention through the Markov process. 
    \textcolor{black}{It avoids the burden caused by non-differentiable operations and further speeds up the approximate attention.}
    Moreover, we extend the proposed model to three downstream tasks including classification, detection, and segmentation. Extensive experiments show the effectiveness of AnchorFormer, e.g., achieving up to a \textit{\textbf{9.0\%}} higher accuracy or \textit{\textbf{46.7\%}} FLOPs reduction on ImageNet classification, \textit{\textbf{81.3\%}} higher mAP on COCO detection under comparable FLOPs, as compared to the current baselines.
\end{abstract}



\end{frontmatter}



\section{Introduction}
\label{sec:intro}

The powerful performance of the Transformers \cite{vaswani2017attention} in the natural language processing has triggered extensive research on Transformers in the computer vision field. As core variants, vision transformers (ViTs) utilize the multi-head self-attention to extract the deep feature representation by partitioning the input images into some patches with identical granularity and processing them as sequences by positional embedding \cite{dosovitskiy2020image}. 
Meanwhile, thanks to capturing the global similarities among the patches, ViTs have achieved excellent performance on various vision tasks, e.g., image classification, object detection, semantic segmentation, andancient text analysis \cite{liu2021swin, AdvCPG, CNN2GNN}.
For instance, in analyzing ancient texts or inscriptions, ViTs can assist with character recognition, even in degraded or stylized scripts.
In particular, the ability of Transformers to process global details can effectively facilitate the recognition of ancient characters.
The vision transformer is also extended to multi-modal data \cite{CLIP}. 
For example, the method proposed by \cite{multiview} is a very interesting and promising framework that learns the underlying relevance among various modalities by a low-dimensional manifold learning mechanism. It can effectively deal with the weak features of some modalities that contain a lot of noise.  

\begin{figure}[t]
    \centering
    \includegraphics[width=0.6\linewidth]{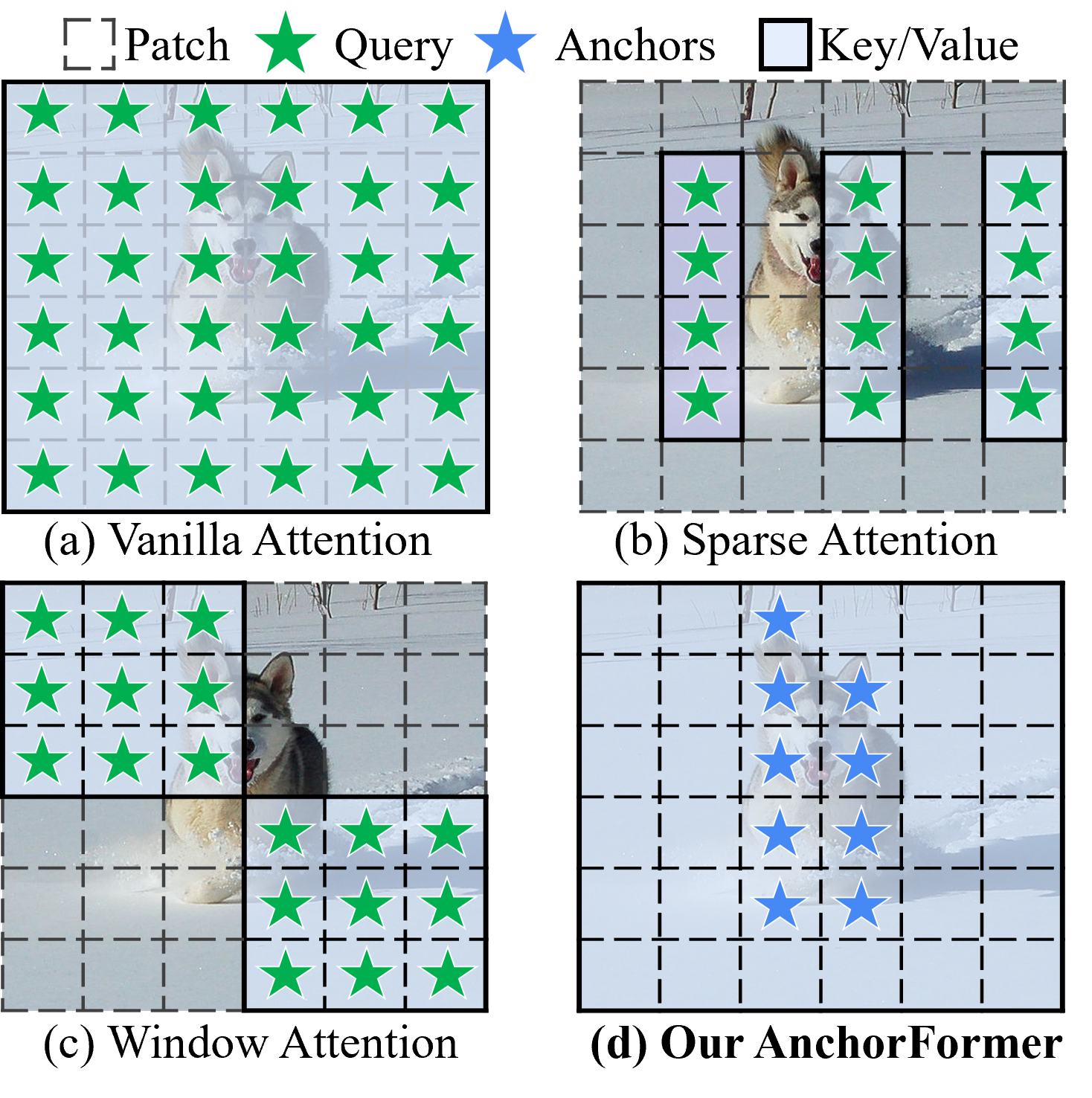}\\
    \caption{Comparison of the proposed and other efficient attention strategies.  Fig. 1a is the vanilla self-attention in ViTs. Fig. 1b is the sparse-based attention which mainly preserves the specific queries, keys, and values. Fig. 1c is the window attention which calculates the local attention within the windows. Fig. 1d is the proposed model which focuses on highly informative regions and differentiable learning the pivotal attention.}
    \label{intro_1}
\end{figure}

While ViTs have shown effectiveness on computer vision tasks, the computation complexity is a main bottleneck to limit development. Specifically, due to estimating the global self-attention by calculating the inner product between each token, the complexity of attention grows quadratically with the number of input tokens, i.e., $\mathcal{O}(n^2)$ \cite{pan2023slide}. 
It leads to excessive computation costs when handling the high-resolution inputs and will be impractical to extend on the limited memory device. To overcome this problem, a promising insight is to introduce sparse attention to ViTs \cite{zhu2023biformer}. 
It mainly limits ViTs to focus on smaller regions, not global input. 
Among them, PVT \cite{wang2021pyramid} introduces sparse attention to select and estimate the similarities among small regions by calculating the inner production among these queries and keys. 
Then, it will share these sparse similarities to each query-key pair and obtain global attention. However, since the informative features are distributed randomly in the input images, the sparse based methods are weak in learning the local features and even discard the informative features. Different from these sparse strategies, some researchers introduce the window attention paradigm to reduce the complexity \cite{liu2021swin, dong2022cswin}. As shown in Fig. \ref{intro_1}, they divide the input tokens into the pre-designed windows paradigm and limit the ViTs calculating the attention and extracting the deep feature within these windows. Nevertheless, window attention will introduce an extra obstacle as cross-window communication. Besides, this window paradigm also restricts the setting of the model structures like how to window shifts.

Instead of the two referred efficient strategies, a natural and effective insight is introducing the anchor tokens to represent the informative regions and learn the global self-attention based on the link between anchors and other tokens. This similar idea has been widely adopted in many fields \cite{zhang2022non, SGNN, redmon2018yolov3}. Among them, \cite{zhang2022non} transmits the graph into a bipartite graph by introducing the anchors, which effectively accelerates the inference of the graph neural networks. \cite{redmon2018yolov3} reduces the computation cost by estimating the IoU between objects and anchors. 
Although an anchor-based strategy can accelerate the model inference, the key concerns for extending it on ViTs are how to select the proper anchors and learn the global similarities from the anchor distributions.

\begin{figure}[t]
    \centering
    \subcaptionbox{\label{comp_classification}Image Classification on ImageNet Dataset}{
        \includegraphics[width=0.9\linewidth]{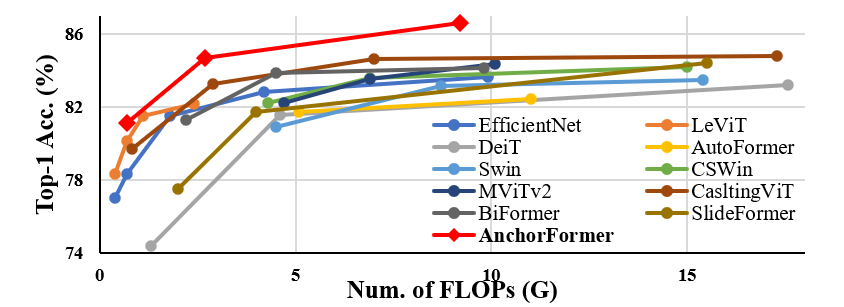}
    }
    \subcaptionbox{\label{comp_detection}Object Detection on Common Objects in Context (COCO) Dataset}{
        \includegraphics[width=0.9\linewidth]{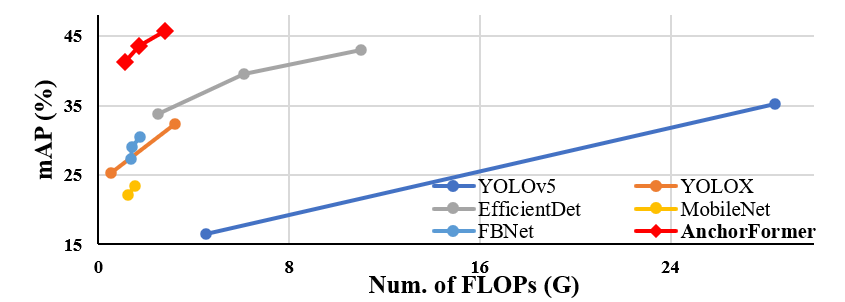}
    }
    \caption{{The proposed AnchorFormer and all baselines. 
    Fig. \ref{comp_classification} is the image classification results on the ImageNet dataset. 
    Fig. \ref{comp_detection} is the object detection results on the COCO dataset.}}
    \label{intro_2}
    \vspace{-5mm}
\end{figure}

In this paper, we propose an anchor-based efficient vision transformer (\textbf{AnchorFormer}) which introduces the anchor tokens to accelerate the ViTs and reduce the complexity from $\mathcal{O}(n^2)$ to $\mathcal{O}(mn)$, where $m$ is an anchor number and $m < n$. Compared with the sparse-based method PVT, the proposed AnchorFormer can differentiable learn the pivotal information among the image datasets. Meanwhile, by the Markov process, the proposed anchor-based strategy can accurately learn the global self-attention from the anchors. Thus, it is flexible to generalize on many ViTs compared with window-based models such as Swin \cite{liu2021swin}. As shown in Fig. \ref{intro_2}, the proposed AnchorFormer has consistently achieved the best trade-off between accuracy and efficiency over the other baselines on image classification and detection tasks. 
Our core contributions are as follows:
\begin{itemize}
    \item To reduce the complexity and accelerate the ViTs, we design an anchor-based method to represent the informative queries. It can generate pivotal attention by estimating the bipartite attention between them and tokens.
    \item To differentiable and accurately learn the pivotal regions distributed randomly in the input images, we represent the anchor tokens with neurons and utilize the neural layer to fit the distribution.
    \item Inspired by the Markov process, the global similarities can be obtained from the distribution of anchors. Meanwhile, by rearranging the multiplication orders, it obtain the linear complexity such as $\mathcal{O}(mn)$. Extensive experiments show the effectiveness of the proposed model, i.e., {{9.0\%}} higher accuracy or {{46.7\%}} FLOPs reduction on classification.
\end{itemize}

{
\color{black}

In the following paper, we introduce the efficient ViTs in Section \ref{section_related_work} 
and then elaborate on the details of the proposed AnchorFormer in Section \ref{section_method}. 
Extensive experimental results are summarized in Section \ref{section_experiments}. 

\section{Related Work} \label{section_related_work}

In this section, we review some works carried out to improve the efficiency and accelerate the transformers. Firstly, some researchers introduce the pruning and decomposition strategies. \cite{michel2019sixteen} proves that all attention heads are not required for specific downstream tasks. It removes some heads and reduces the model parameters by estimating the influence of each head on the final output. Meanwhile, some works attempt to reduce the depths instead of the width of transformers \cite{fan2019reducing, hou2020dynabert}. Apart from the pruning, matrix decomposition is also employed to improve the efficiency \cite{wang2019structured}. 
Further, several studies draw inspiration from the Transformer strategy while ingeniously avoiding the high complexity of the architecture.
Wang et al. \cite{WZ1} propose a lightweight encoder that aggregates token representations via semantic correlation and tensor decomposition. This clever aggregation surpasses standard Transformers on text-classification benchmarks while operating at a substantially reduced computational cost.
In parallel, Hou et al. \cite{WZ2} introduces an effective graph-temporal encoder that dynamically weights time-step signals via a novel temporal mechanism and seamlessly fuses them with graph topology. This elegant design unlocks richer structural and temporal representations while maintaining a markedly lower cost than full Transformer variants.
Cheng et al. \cite{WZ3} develops a novel positional-encoding scheme that can capture and embed the sequence order in an efficient, lightweight fashion.
Secondly, knowledge distillation also is utilized to improve efficiency. Mukherjee et al. \cite{mukherjee2020xtremedistil} utilizes a pre-trained BERT model as the teacher to guide the student transformer training. For vision transformers, \cite{jiaefficient} introduces the manifold learning into the distillation to explore the relationship among the patches and improve performance. Thirdly, there is a lot of work on how to introduce quantization in transformers \cite{bhandare2019efficient, liu2021post}. Literature \cite{shridhar2020end} represents the input with binary high-dimensional vectors and reduces the complexity. \cite{prato2019fully} proposes a fully quantized transformer to handle the machine translation tasks.
Gao et al. \cite{WZ4} propose an innovative Fourier Transformer that performs spectral regularization directly in the frequency domain, thereby elegantly mitigating spatial-domain redundancy and noise in the data.
Lastly, more researchers pay attention to designing  compact transformer architecture. Our model also belongs to this category. The neural architecture search (NAS) is introduced to automatically search for the best compact architecture \cite{guo2019nat, so2019evolved}. Inspired by the graph theory, some models introduce sparsity in estimating the similarity among the tokens \cite{wang2021pyramid, spielman2011spectral, zaheer2020big}. However, these methods may discard some informative features due to constructing the sparse attention. Although the slide windows based strategy can solve this problem and reduce the complexity simultaneously \cite{liu2021swin, dong2022cswin}, the window attention introduces an extra obstacle to cross-window communication. 
}

\section{AnchorFormer} \label{section_method}
\begin{figure}[t]
    \centering
    \includegraphics[width=0.9\linewidth]{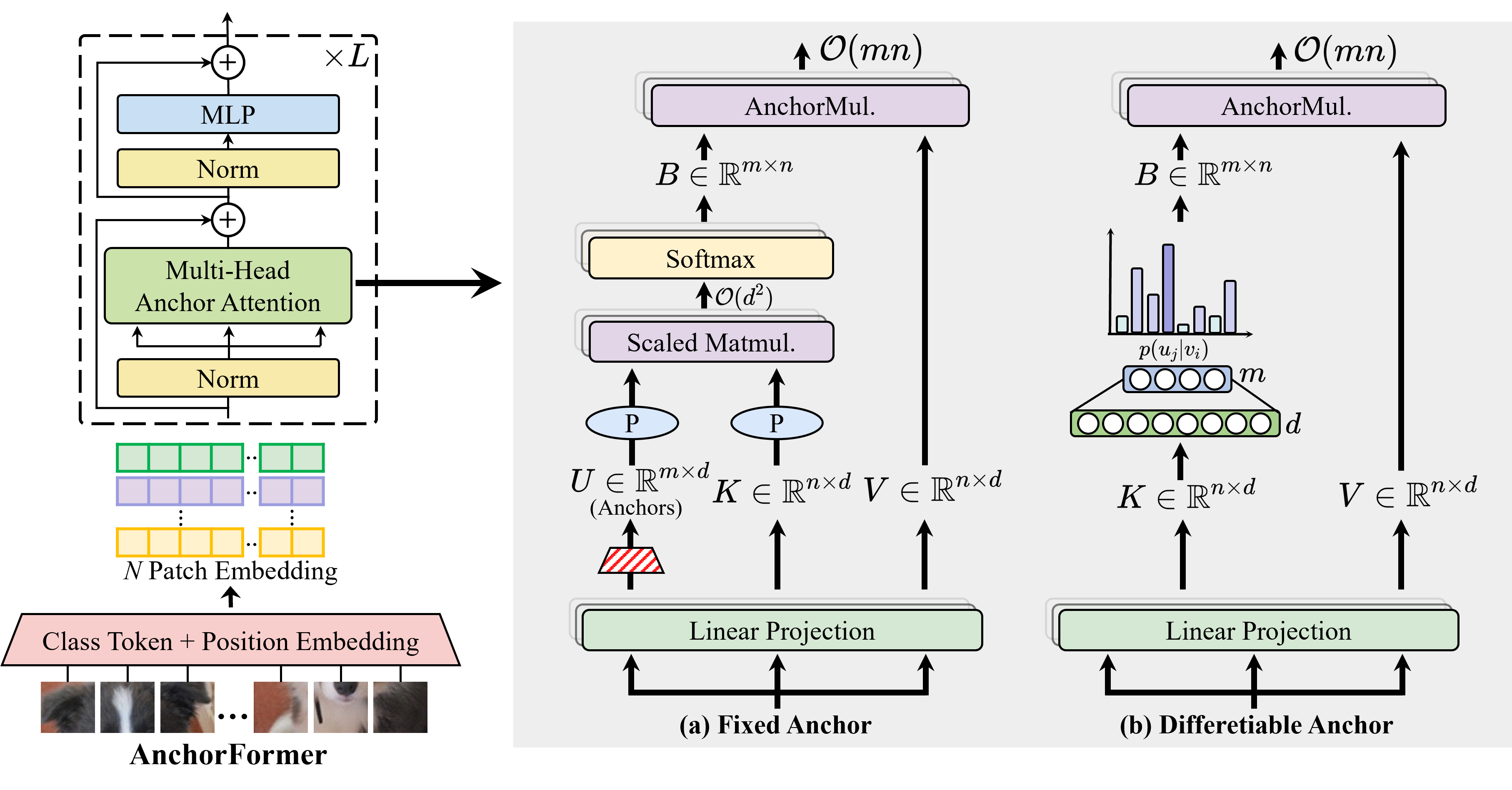}\\
    \caption{{The architecture of the proposed AnchorFormer.} AnchorFormer has $L$ multi-head anchor attention layers. It splits the input image into $n$ patches. Fig. (a) is the proposed basic anchor-based attention layer. It selects $m$ anchors among the queries $\bm Q$ and calculates the pivotal similarities. Fig. (b) utilizes a deep neural layer to represent $m$ anchors and differentiable learn the pivotal similarities. The block is named \textbf{AnchorMul.} can learn the global attention based on the Markov process.}
    \label{framework_fig}
\end{figure}

We propose an anchor-based vision transformer (\textbf{AnchorFormer}) in this section. It can reduce the complexity from $\mathcal{O}(n^2)$ to $\mathcal{O}(mn)$ by estimating the bipartite attention between anchors and other tokens, where $m$ is the number of tokens. 
The framework is illustrated in Fig. \ref{framework_fig}.

\subsection{Motivation}
Recently, vision transformers (ViTs) have shown impressive performance on vision tasks. 
As a core component of ViTs, the self-attention module generally consists of multiple heads \cite{you2023castling}. Given $n$ tokens, each head can capture the global information by measuring the similarities among all tokens as
\begin{equation}
    \label{self_attention}
    \bm h_t^m= \sum_{i=1}^n \frac{{\rm exp} (\bm q_t \bm k_i^T / \sqrt{d} )}{ \sum_{j=1}^n {\rm exp} (\bm q_t \bm k_j^T / \sqrt{d} )} \bm v_i,
\end{equation}
where $\bm q_t \in \bm Q$, $\bm k_i \in \bm K$, $\bm v_i \in \bm V$, $\bm h_t \in \bm H$ are the row vectors, ${\rm exp}(\cdot)$ is an exponential function, and $m$ is the $m$-th head. $\bm H \in \mathbb{R}^{n \times d}$ is an attention matrix. $\bm Q$, $\bm K$, $\bm V  \in \mathbb{R}^{n \times d}$ are the query, key, and value. They are both obtained by projecting the tokens $\bm X \in \mathbb{R}^{n \times D}$ with three learnable weights $\bm W^Q$, $\bm W^K$, $\bm W^V \in \mathbb{R}^{D \times d}$. Eq. (\ref{self_attention}) estimates the similarities between each pair of tokens by calculating the inner product between the query-key pairs, which has $\mathcal{O}(n^2)$ complexity and takes the vast costs. Besides, enlightened that the pivotal information of the input image often randomly gathers in a few regions, the model could pay more attention to the similarities among these regions. To sum up, there is a natural question, \textbf{\textit{how to efficiently accelerate the ViTs for learning the pivotal similarities?}}

\subsection{Accelerate ViTs with Anchor Tokens}

In this work, the distributions among the tokens are denoted by the conditional probability such as $p(\bm v_j | \bm v_i)$. Correspondingly, the similarities between them can be viewed as the sampling results from $p(\bm v_j | \bm v_i)$. Thus, the global similarities learned by vanilla ViTs is reformulated as
\begin{equation}
    \label{p_sim}
    p(\bm v_j | \bm v_i)= \frac{{\rm exp} (\bm q_j \bm k_i^T / \sqrt{d} )}{ \sum_{j=1}^n {\rm exp} (\bm q_j \bm k_i^T / \sqrt{d} )},
\end{equation} 
where $\sum_{j=1}^n p(\bm v_j | \bm v_i)=1$ and $p(\bm v_j | \bm v_i) = p(\bm v_i | \bm v_j)$. As shown in Eq. (\ref{p_sim}), vanilla ViTs mainly compute the inner products between the query $\bm Q$ and the key $\bm K$ to measure the global self-attention. Thus, to accelerate the ViTs, a direct insight is selecting some representative tokens named \textbf{\textit{anchors}} $\bm U \in \mathbb{R}^{m \times d}$, where $m$ is the number of anchors \cite{zhang2022non}. Then, to obtain the global self-attention $\bm H$ and accelerate the ViTs, we not only need to obtain the pivotal distributions as $p(\bm v_j | \bm u_i)$ but also attempt to estimate the global distributions $p(\bm v_j | \bm v_i)$ from them. 

Specifically, the pivotal similarities between anchors and tokens can be viewed as sampling results from $p(\bm u_j | \bm v_i)$. Meanwhile, since the anchors indicate the more representative tokens, the ideal anchors should satisfy the following problem as
\begin{equation}
    \label{anchors_problem}
    \min_{\bm u, p(\cdot|\bm v_i)} \sum_{i=1}^{n} \mathbb{E}_{\bm u \sim p(\cdot|\bm v_i)} {\rm dist} (\bm u, \bm v_i),
\end{equation}
where ${\rm dist}(\bm u, \bm v_i)$ mainly measures the distance between the anchors and other tokens. Notably, we take the same inner product and normalized strategy as vanilla ViTs to obtain $p(\bm u_j|\bm v_i)$,
\begin{equation}
    \label{p_distribution}
    p(\bm u_j|\bm v_i) = \frac{{\rm exp}(\bm u_j \bm k_i^T / \sqrt{d})}{\sum_{j=1}^{m}{\rm exp}(\bm u_j \bm k_i^T / \sqrt{d})},
\end{equation}
where $p(\bm u_j | \bm v_i) = p(\bm v_i | \bm u_j)$. Meanwhile, following vanilla ViTs, we also introduce $\bm k_i$ to measure the relationship between the tokens. Correspondingly, Eq. (\ref{anchors_problem}) is reformulated as 
$
    \min_{\bm u, p(\cdot|\bm v_i)} \sum_{i=1}^{n} \mathbb{E}_{\bm u \sim p(\cdot|\bm v_i)} \|\bm u - \bm k_i\|_2^2.
$
Then, the anchors can be solved by taking its derivative regarding $\bm u$ and setting it to $0$,
\begin{equation}
    \label{anchors}
    \bm u = \frac{\sum_{i=1}^n p(\bm u|\bm v_i) \bm k_i}{\sum_{i=1}^n p(\bm u|\bm v_i)}.
\end{equation}
To estimate the bipartite attention between anchors and all tokens, we further reformulate $p(\bm u_j | \bm v_i)$ and $p(\bm v_i | \bm u_j)$ by matrix form. 
Define $\bm A \in \mathbb{R}^{n \times m}$ as $a_{ij}=p(\bm u_j | \bm v_i)$ and 
\begin{equation}
    \bm G=\begin{bmatrix}
        \bm 0 & \bm A \\
        \bm A^T & \bm 0
    \end{bmatrix},
    \bm D=\begin{bmatrix}
        \bm I & \bm 0 \\
        \bm 0 & \bm \Delta
    \end{bmatrix},
\end{equation}
where $\bm D$ is diagonal and $d_{ii}=\sum_{j=1}^{n+m} t_{ij}$. To bridge the anchors and tokens, we introduce a probability transferring matrix, 
\begin{equation}
    \label{transfer_matrix}
    \bm F = \bm D^{-1} \bm G=
    \begin{bmatrix}
        \bm 0 & \bm A \\
        \bm \Delta^{-1} \bm A^T & \bm 0
    \end{bmatrix}
    \in \mathbb{R}^{(n+m) \times (n+m)}.
\end{equation}
Then, according to the Markov process, $p(\bm v_j | \bm v_i)$ can be estimated by the one-step transition probability as
\begin{equation}
    \label{p_kk}
    p(\bm v_j | \bm v_i) = \sum_{l=1}^{m} p(\bm v_j | \bm u_l) p(\bm u_l | \bm v_i).
\end{equation}
Similarly, $p(\bm u_j | \bm u_i)=\sum_{l=1}^{n} p(\bm u_j | \bm v_l) p(\bm v_l | \bm u_i)$. Thus, the one-step transition probability is formulated as
\begin{equation}
    \label{one_step}
    \bm F^2=\begin{bmatrix}
        \bm A \bm \Delta^{-1} \bm A^T & \bm 0 \\
        \bm 0 & \bm \Delta^{-1} \bm A^T \bm A
    \end{bmatrix}
    =\begin{bmatrix}
        \bm S_t & \bm 0 \\
        \bm 0 & \bm S_u
    \end{bmatrix},
\end{equation}
where $\bm S_t$ indicates the self-attention among $n$ tokens, constructed by $m$ anchors, and $\bm S_u$ shows the similarities between the anchors. Besides, $\bm S_t$ has been normalized because of
\begin{equation}
    \label{normalized_S}
    \bm S_t \bm 1_n = \bm A \bm \Delta^{-1} \bm A^T \bm 1_n = \bm 1_n,
\end{equation}
where $\bm 1_n \in \mathbb{R}^{n \times 1}$ \textcolor{black}{denotes a vector with all entries as 1}. Thus, the global similarities among the tokens are the sampling results of $\bm S_t$ and the global self-attention $\bm H$ is calculated by
\begin{equation}
    \label{attention_H}
    \bm H = \bm A \bm \Delta^{-1} \bm A^T \bm V.
\end{equation}
Then, we explain how Eq. (\ref{attention_H}) accelerates the ViTs. It should be emphasized that $\bm S_t$ cannot be calculated explicitly. The core idea is to \textbf{\textit{rearrange the multiplication order}}, 
\begin{equation}
    \label{anchor_H}
    \bm M_1 = \bm B^T \bm V \Rightarrow \bm M_2 = \bm \Delta^{-1} \bm M_1 \Rightarrow \bm M_3 = \bm B \bm M_2,
\end{equation}
where the complexity of \textcolor{black}{calculating} $\bm M_1$, $\bm M_2$ and $\bm M_3$ are $\mathcal{O}(mnd)$, $\mathcal{O}(m^2d)$, and $\mathcal{O}(nmd)$, respectively. Thus, the computational complexity of Eq. (\ref{anchor_H}) is $\mathcal{O}(nmd + m^2d)$. Because $m$ and $d$ generally are smaller than $n$, the complexity is reduced to $\mathcal{O}(nm)$. More importantly, if the number of anchors is small enough, we just take $\mathcal{O}(n)$ to obtain global self-attention. 

\subsection{AnchorFormer: Differentiable Anchor ViTs}
Since the pivotal information is often randomly gathered in the input images, an ideal strategy is utilizing the neural layer to fit their distributions and differentiable learn the pivotal similarities $\bm B$. 
Meanwhile, indicated by Eq. (\ref{anchor_H}), the global self-attention $\bm H$ directly depends on the pivotal similarities. Thus, we design an the AnchorFormer that focuses on differentiable learning the pivotal similarities in attention heads.

Specifically, we generate the keys $\bm K$ and the values $\bm V$ by two learnable parameters $\bm W^{K}$ and $\bm W^{V}$, respectively. Notably, to extract the deep information, a neural layer generally in a deep neural network (DNN) introduces a learnable projection matrix and calculates the inner product between it and the input data. Inspired by it, Eq. (\ref{p_distribution}) can be fitted with a neural layer as
\begin{equation}
    \label{anchor_p}
    p(\bm u_j|\bm v_i) = \frac{{\rm exp}(\bm w^{S}_j \bm k_i^T / \sqrt{d})}{\sum_{j=1}^{n}{\rm exp}(\bm w^{S}_j \bm k_i^T / \sqrt{d})},
\end{equation}
where $\bm w^{S}_j \in \mathbb{R}^{m \times d}$ is an anchor and can be implemented by learnable parameters. Thus, the pivotal similarities $p(\bm u_j|\bm v_i)$ can be learned differentiable and the learnable anchors can accurately mine the latent distribution of the pivotal region for whole input images by gradient descent. Since the proposed anchor-based attention can be calculated independently, it can naturally be extended to multi-head self-attention learning. Among them, each head can capture the global self-attention by Eq. (\ref{anchor_p}) and Eq. (\ref{anchor_H}). Then, for $l$ heads, the global multi-head self-attention is calculated by 
$
    \bm H^{M} = (\bm H^1 \oplus \bm H^2 \oplus ... \oplus \bm H^l) \bm W
$,
where $\bm W \in \mathbb{R}^{ld \times d}$ is a projection matrix and $\oplus$ is the matrix concatenation operation.


\section{Experiments} \label{section_experiments}

\subsection{Experimental Settings} 
\textbf{Datasets and Tasks:} To verify the performance and efficiency of the proposed model, we introduce three representative datasets and three mainstream computer vision tasks, i.e., image classification on the ImageNet dataset \cite{deng2009imagenet}, object detection on the COCO dataset \cite{lin2014microsoft}, and semantic segmentation on ADE210K \cite{zhou2017scene}. 

\begin{table}[t]
    \renewcommand\arraystretch{1.2}
    \small
    \centering
    \caption{Classification on ImageNet Dataset. The baselines are divided into three categories according to the FLOPs. The last column represents the Top-1 improvements of each model than three baselines including LeViT-192, DeiT-S and DeiT-B. }
    \vspace{-2mm}
    \scalebox{0.70}{
        \begin{tabular}{p{1.4cm}<{\centering} | p{3.5cm}<{\raggedright} | p{1.3cm}<{\centering} p{1.3cm}<{\centering} | p{1.3cm}<{\centering} p{1.3cm}<{\centering}}
            \toprule
            \textbf{\begin{tabular}[c]{@{}c@{}}FLOPs\\ Ranges\end{tabular}} 
            & \textbf{\begin{tabular}[c]{@{}c@{}}Models \end{tabular}} 
            & \textbf{\begin{tabular}[c]{@{}c@{}}Params\\ (M)\end{tabular}} 
            & \textbf{\begin{tabular}[c]{@{}c@{}}FLOPs\\ (G)\end{tabular}} 
            & \textbf{\begin{tabular}[c]{@{}c@{}}Top-1\\ Acc.(\%)\end{tabular}} 
            & \textbf{\begin{tabular}[c]{@{}c@{}}Top-5\\ Acc.(\%)\end{tabular}} \\
            \midrule
            \multirow{6}{*}{\textless{}1G} & EfficientNet-B0              & 5.3                 & 0.4              & 77.05              & \underline{91.27}              \\
            & EfficientNet-B1              & 7.8                 & 0.7              & 78.32              & 90.72   \\
            & LeViT-128                    & 9.2                 & 0.4              & 78.33              & 89.18   \\
            & LeViT-192                    & 10.9                & 0.7              & \underline{80.13}              & 90.53   \\
            & CastlingViT-192              & 12.7                & 0.8             & 79.73              & \underline{91.27}   \\
            & \cellcolor[RGB]{232,240,254}\textbf{AnchorFormer-T} & \cellcolor[RGB]{232,240,254}4.2                 & \cellcolor[RGB]{232,240,254}0.7              & \cellcolor[RGB]{232,240,254}\textbf{81.12}     & \cellcolor[RGB]{232,240,254}\textbf{93.86}   \\ \midrule
            \multirow{16}{*}{1$\sim$5G}            
            & Sparsifiner                   & 4.2                  & 1.3              & 74.38              & 91.97   \\
            & Combiner                   & 6.4                  & 1.5              & 68.37              & 76.81   \\
            & Sparsifiner                   & 5.3                  & 1.8              & 64.72              & 73.74   \\
            & EfficientNet-B3              & 12.3                  & 1.8              & 81.54              & 94.85  \\
            & EfficientNet-B4              & 19.1                  & 4.2              & 82.83              & 96.43   \\
            & DeiT-T                       & 5.6                 & 1.3              & 74.38              & 91.97   \\
            & DeiT-S                       & 21.9                & 4.6              & 81.58              & 95.06   \\
            & LeViT-256                    & 18.9                & 1.1              & 81.53              & 94.17   \\
            & LeViT-384                    & 39.1                & 2.4              & 82.17              & 95.36   \\
            & Swin-T                       & 29.6                  & 4.5              & 80.94              & 94.33   \\
            & CSWin-T                      & 23.4                  & 4.3              & 82.22              & 93.46   \\
            & PVTv2-V2                     & 25.5                  & 4.2                & 81.86              & 93.65   \\
            & MViTv2-T                     & 24.1                  & 4.7              & 82.21              & 94.16   \\
            & CastlingViT-384              & 45.8                & 2.9              & 83.26              & \underline{96.69}   \\
            & SlideFormer-T                & 12.2                & 2.6                & 77.51              & 91.67   \\
            & SlideFormer-S                & 22.7                & 4.5                & 81.75              & 92.48   \\
            & BiFormer-T                   & 13.1                & 2.2              & 81.32              & 93.31   \\
            & BiFormer-S                   & 26.4                  & 4.5              & \underline{83.89}              & 93.86   \\
            & \cellcolor[RGB]{232,240,254}\textbf{AnchorFormer-S} & \cellcolor[RGB]{232,240,254}18.6                & \cellcolor[RGB]{232,240,254}2.7              & \cellcolor[RGB]{232,240,254}\textbf{84.69}     & \cellcolor[RGB]{232,240,254}\textbf{96.82}    \\ \midrule
            \multirow{15}{*}{$>$5G}      & EfficientNet-B5              & 30.7                  & 9.9              & 83.64              & 96.13      \\
            & DeiT-B                       & 86.3                & 17.5            & 83.21              & 96.13     \\
            & Swin-S                       & 50.3                  & 8.7              & 83.16              & \underline{96.17}     \\
            & Swin-B                       & 88.1                  & 15.4             & 83.51              & 96.14     \\
            & CSWin-S                      & 35.6                  & 6.9              & 83.57              & 95.75     \\
            & CSWin-B                      & 78.3                  & 15.2               & 84.22              & 96.14     \\
            & AutoFormer-S                 & 22.9                & 5.1              & 81.72              & 95.73     \\
            & AutoFormer-B                 & 54.8                  & 11.7               & 82.47              & 95.64     \\
            & MViTv2-S                     & 34.7                & 6.9              & 83.56              & 94.71     \\
            & MViTv2-B                     & 51.2                & 10.1             & 84.37              & 95.06     \\
            & CastlingViT-S                & 34.7                & 7.3                & 84.63              & 95.36     \\
            & CastlingViT-B                & 87.2               & 17.3             & \underline{84.82}              & 95.87     \\
            & SlideFormer-B                & 89.7                  & 15.5             & 84.42              & 96.07     \\
            & BiFormer-B                   & 58.3                  & 9.8              & 84.13              & 95.88     \\
            & \cellcolor[RGB]{232,240,254}\textbf{AnchorFormer-B} & \cellcolor[RGB]{232,240,254}63.5                & \cellcolor[RGB]{232,240,254}9.2              & \cellcolor[RGB]{232,240,254}\textbf{86.62}     & \cellcolor[RGB]{232,240,254}\textbf{97.84}    \\ \bottomrule
    \end{tabular}
    }
    
    \vspace{-5mm}
    \label{classification_result}
\end{table}

\begin{figure}
    \centering
    \includegraphics[width=\linewidth]{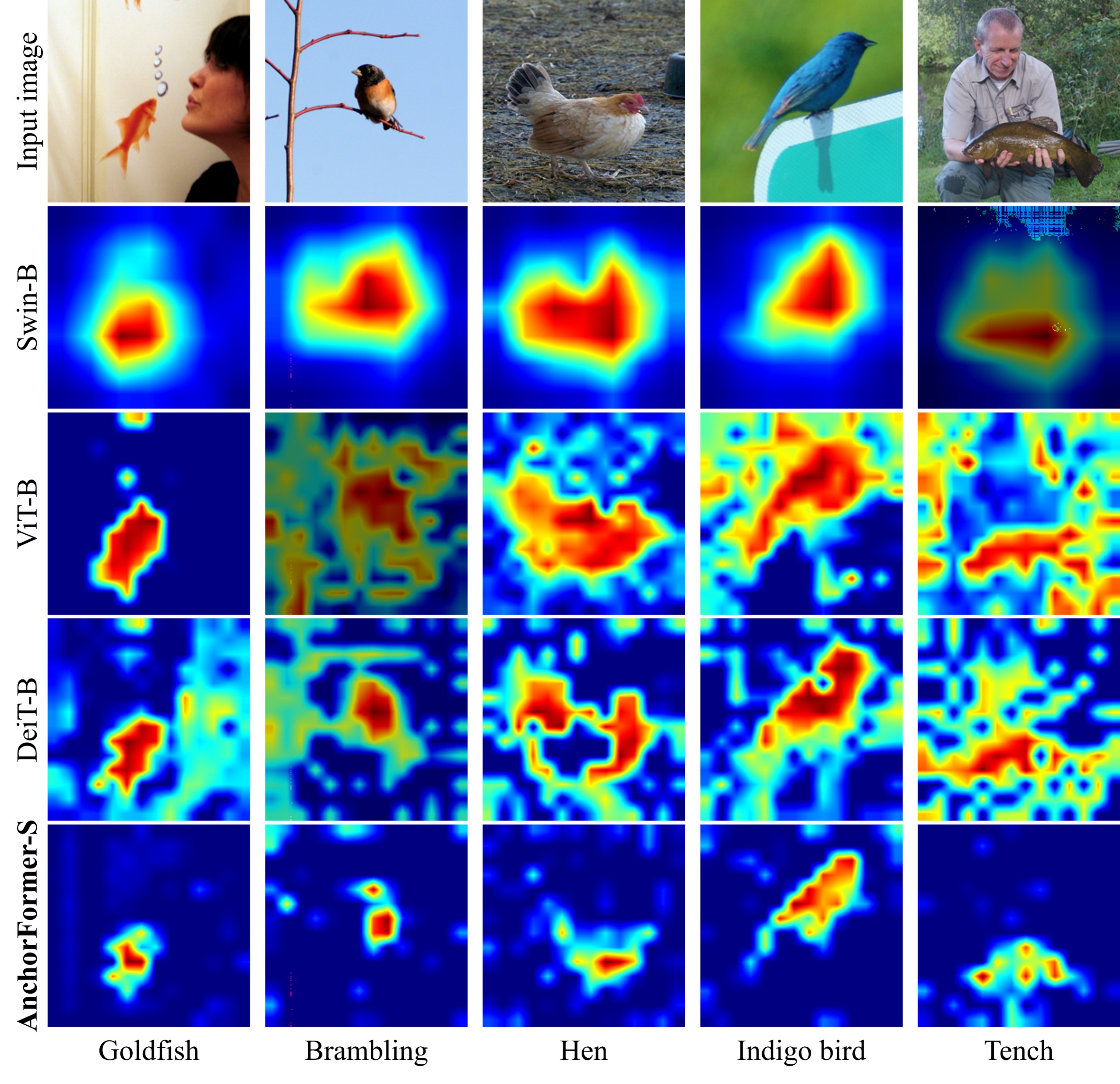}\\
    \vspace{-2mm}
    \caption{Visual explanations generated by different models on the ImageNet validation dataset \cite{deng2009imagenet}. From top to down: input image, visual explanation maps of the Swin-B \cite{liu2021swin}, ViT-B, DeiT-B \cite{touvron2021training} and AnchorFormer-S, respectively.}
    \vspace{-3mm}
    \label{visualize}
\end{figure}

\begin{table}[t]
    \renewcommand\arraystretch{1.2}
    \centering
    \caption{AnchorFormer over SOTA detection baselines on the COCO dataset, where the proposed model replaces the last stage of ESNet. ``-T/S/B'' and ``-512/640/768'' mean tiny/small/base and different input resolution, respectively.}
    \vspace{-1mm}
    \scalebox{0.73}{
        \begin{tabular}
            {p{3.3cm}<{\raggedright} | p{1.1cm}<{\centering}  p{1.1cm}<{\centering} | p{1.1cm}<{\centering}  p{1.1cm}<{\centering} p{1.1cm}<{\centering}}
            \toprule
            \textbf{\begin{tabular}[c]{@{}c@{}}Models \end{tabular}} 
            & \textbf{\begin{tabular}[c]{@{}c@{}}Params\\ (M)\end{tabular}} 
            & \textbf{\begin{tabular}[c]{@{}c@{}}FLOPs\\ (G)\end{tabular}}  
            & \textbf{\begin{tabular}[c]{@{}c@{}}mAP\end{tabular}} 
            & \textbf{\begin{tabular}[c]{@{}c@{}}AP50\end{tabular}} 
            & \textbf{\begin{tabular}[c]{@{}c@{}}AP75\end{tabular}}
            \\
            \midrule
            YOLOv5-N              & 1.9                 & 4.5              & 28.38              & 45.21     & - \\
            YOLOv5-S              & 7.2                 & 16.5              & 35.18              & 52.55        & - \\
            YOLOX-Nano            & 0.91                & 0.54              & 25.23              & -        & -     \\
            YOLOX-Tiny            & 5.06                & 3.23              & 32.37              & -        & -     \\
            \midrule
            EfficientDet-512              & 3.9                 & 2.5              & 33.81              & 52.24     & 35.82 \\
            EfficientDet-640              & 6.6                 & 6.1              & 39.58              & 58.59     & 42.28 \\
            EfficientDet-768            & 8.1                & 10.8              & \underline{43.25}              & \underline{62.31}       & \underline{46.18} \\
            \midrule
            MobileNetV2              & 4.3                 & 1.2              & 22.13              & -      & - \\
            MobileNetV3            & 3.3                & 1.5              & 23.48              & -     & - \\
            \midrule
            FBNetV5-$A$            & -                 & 1.4              & 27.35              & -      & - \\
            FBNetV5-$A_C$           & -                & 1.4              & 28.98              & -      & - \\
            FBNetV5-$A_R$            & -                & 1.8              & 30.52              & -     & - \\
            \midrule
            CastlingViT-S   & {3.3}                  & {1.1}             & 31.34              & 46.69       & 32.52  \\
            CastlingViT-M   & 6.0                & 2.0             & 34.07              & 49.63     & 35.83  \\
            CastlingViT-L   & 13.1                & 5.3             & 34.42              & 53.59        & 39.58  \\
            \midrule
            \cellcolor[RGB]{232,240,254}\textbf{AnchorFormer-T}            & \cellcolor[RGB]{232,240,254}\textbf{2.9}                & \cellcolor[RGB]{232,240,254}\textbf{1.1}              & \cellcolor[RGB]{232,240,254}\textbf{41.23}              & \cellcolor[RGB]{232,240,254}\textbf{60.57}     & \cellcolor[RGB]{232,240,254}\textbf{45.46}      \\
            \cellcolor[RGB]{232,240,254}\textbf{AnchorFormer-S}            & \cellcolor[RGB]{232,240,254}\textbf{5.5}                & \cellcolor[RGB]{232,240,254}\textbf{1.7}              & \cellcolor[RGB]{232,240,254}\textbf{43.61}              & \cellcolor[RGB]{232,240,254}\textbf{62.83}     & \cellcolor[RGB]{232,240,254}\textbf{48.29}       \\
            \cellcolor[RGB]{232,240,254}\textbf{AnchorFormer-B}            & \cellcolor[RGB]{232,240,254}\textbf{8.6}                & \cellcolor[RGB]{232,240,254}\textbf{2.8}              & \cellcolor[RGB]{232,240,254}\textbf{45.75}              & \cellcolor[RGB]{232,240,254}\textbf{65.35}     & \cellcolor[RGB]{232,240,254}\textbf{51.81}       \\
            \bottomrule
    \end{tabular}}
    
    \label{Object_detect_result}
\end{table}

\begin{figure}[t]
    \centering
    \subcaptionbox{\label{comp_deit}Comparison on DeiT}{
        \includegraphics[width=0.46\linewidth]{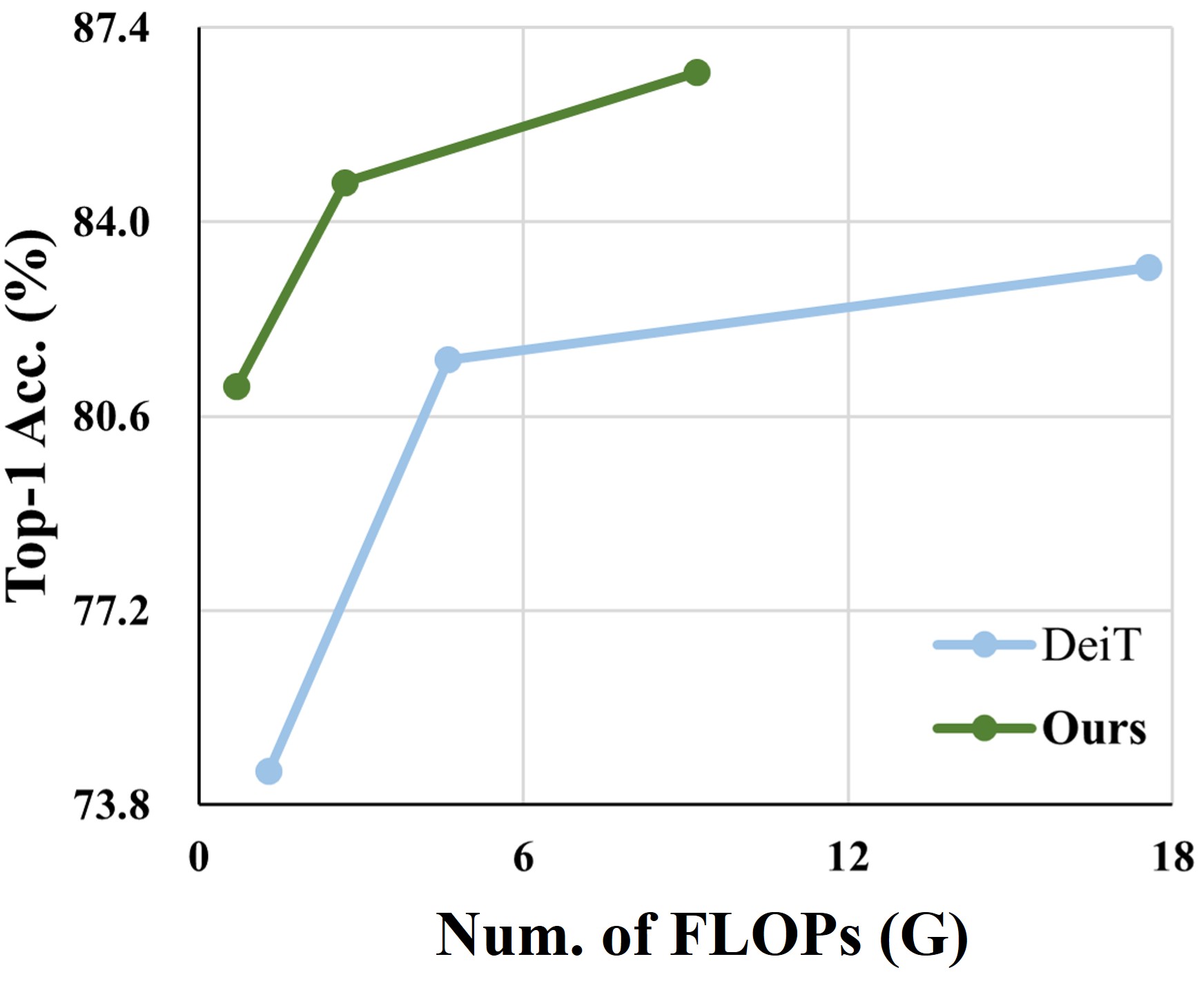}
    }
    \subcaptionbox{\label{comp_levit}Comparison on LeViT}{
        \includegraphics[width=0.46\linewidth]{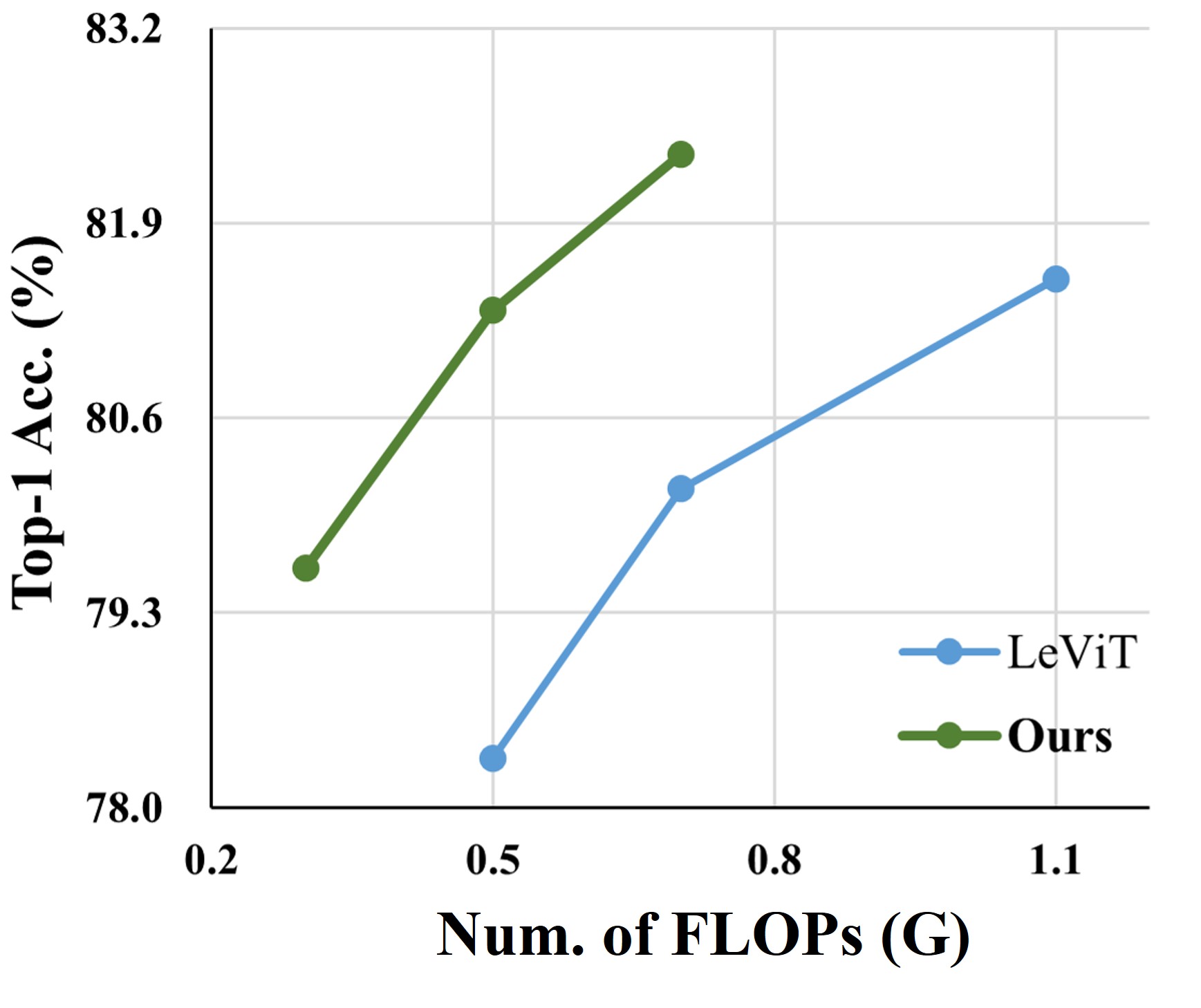}
    }
    \caption{{AnchorFormer vs. baseline on ImageNet.} Fig. \ref{comp_deit} shows the achievement than DeiT and Fig. \ref{comp_levit} shows the improvements than LeViT.}
    \label{Comp_res}
    \vspace{-5mm}
\end{figure}

\textbf{Comparative Methods}
For the classification, we introduce several current efficient deep models as the baseline. It includes 
EfficientNet \cite{tan2019efficientnet}, LeViT \cite{graham2021levit}, DeiT \cite{touvron2021training}, AutoFormer \cite{chen2021autoformer}, Swin \cite{liu2021swin}, CSWin \cite{dong2022cswin}, MViTv2 \cite{li2022mvitv2}, PVT \cite{wang2021pyramid}, CastlingViT \cite{you2023castling},  BiFormer \cite{zhu2023biformer} and SlideFormer \cite{pan2023slide}. Besides, three sparse attention based ViT, Sparsifiner \cite{Sparsifiner}, Combiner \cite{Combiner} and ClusterFormer \cite{ClusterFormer}, are also employed as baselines.
For object detection, five famous detection models are employed as the baselines, i.e., YOLOv5, YOLOX \cite{ge2021yolox}, EfficientDet \cite{tan2020efficientdet}, MobileDet \cite{xiong2021mobiledets}, FBNetV5 \cite{wu2021fbnetv5} and the current CastlingViT.
For the semantic segmentation, we use ResNet \cite{ResNet}, PVT \cite{wang2021pyramid}, Swin \cite{liu2021swin}, DAT \cite{DAT}, Focal \cite{FocalViT}, RMT \cite{RMTViT} and GraftViT \cite{GraftViT} as baselines.


\subsection{Visualization and analysis}
We utilize the Grad-CAM \cite{selvaraju2017grad} to visualize and show the deep feature maps of the proposed method and some representative baselines including Swin \cite{liu2021swin}, vanilla ViT, and DeiT. Fig. \ref{visualize} suggests the visual comparison based on randomly chosen images from the ImageNet validation. As can be shown from the figures, these models can make the higher attention on the target area corresponding the category. Compared with the Swin, the ViT-based models can notice the outline or shape of the target category. More importantly, thanks to differentiable learning of the pivotal information with the anchor tokens, the proposed AnchorFormer can distinguish and pay more attention to the target area from the global receptive field, which  significantly improves performance. Notably, according to Table \ref{classification_result}, the parameters and FLOPs of AnchorFormer-S are much lower than the comparative baselines including Swin-B, ViT-B and DeiT-B. It proves that the anchor mechanism can alleviate some redundant parameters to improve the performance and reduce the model size, simultaneously. 

\subsection{Image Classification}
To evaluate the performance on the classification task, we extend AnchorFormer on the representative vision transformer architecture as DeiT \cite{touvron2021training} with different sizes and compare their performance over some current baselines on the ImageNet dataset. The results are listed in Table \ref{classification_result}. For clarity, all methods are divided into three categories according to the FLOPs range: $<1$G, $1\sim 5$G, and $>$5G. Across the different FLOPs ranges, AnchorFormer consistently achieves the best performance than the others in terms of accuracy and efficiency. For instance, AnchorFormer-B obtains $86.62\%$ top-1 accuracy with $9.2$G FLOPs while the current BiFormer-B takes $15.5$G FLOPs to achieve $84.42\%$ accuracy, i.e., $\downarrow\bm{40.6\%}$ FLOPs and $\uparrow \bm{2.6\%}$ accuracy. 
In a word, the proposed model has improved the top-1 accuracy of $\uparrow \bm{1.2\%} \sim \uparrow \bm{5.3\%}$, $\uparrow \bm{1.0\%} \sim \uparrow \bm{13.9\%}$, and $\uparrow \bm{2.1\%} \sim \uparrow \bm{6.0\%}$ than the other baselines under $<1$G FLOPs, $1\sim 5$G FLOPs, and $>$5G FLOPs, respectively. Furthermore, we also introduce the comparative experiments on the apple-to-apple benchmark including, DeiT vs. AnchorDeiT, and LeViT vs. AnchorLeViT. In Fig. \ref{Comp_res}, compared with DeiT, ours achieves $\downarrow\bm{41.3\%} \sim \downarrow\bm{46.7\%}$ FLOPs reductions and $\uparrow \bm{3.8\%} \sim \uparrow \bm{9.0\%}$ better accuracy. Compared with LeViT, AnchorFormer obtains $\downarrow\bm{28.5\%} \sim \downarrow\bm{40.0\%}$ FLOPs reductions and $\uparrow \bm{1.0\%} \sim \uparrow \bm{1.6\%}$ better accuracy.

\begin{table}[t]
    \renewcommand\arraystretch{1.3}
    \centering
    \caption{AnchorFormer over segmentation baselines on the ADE210K dataset, where we extend the proposed AnchorFormer on two segmentation methods, Semantic-FPN and UperNet. The resolution of the input image is $512 \times 2048$.}
    \vspace{-1mm}
    \scalebox{0.76}{
        \begin{tabular}{l|c|cc|cc}
            \toprule
            \textbf{\begin{tabular}[c]{@{}c@{}}Backbone \end{tabular}} 
            & \textbf{\begin{tabular}[c]{@{}c@{}}Method \end{tabular}} 
            & \textbf{\begin{tabular}[c]{@{}c@{}}Params\\ (M)\end{tabular}} 
            & \textbf{\begin{tabular}[c]{@{}c@{}}FLOPs\\ (G)\end{tabular}} 
            & \textbf{\begin{tabular}[c]{@{}c@{}}mIoU\\ (\%)\end{tabular}} 
            & \textbf{\begin{tabular}[c]{@{}c@{}}mACC\\ (\%)\end{tabular}} 
            \\
            \midrule
            ResNet50 & Semantic-FPN & 28.5 & 183.2 & 36.75 & 43.88 \\
            PVT & Semantic-FPN & 29.5 & 221.6 & 42.83 & 52.15 \\
            Swin & Semantic-FPN & 31.9 & 182.3 & 41.56 & 51.42 \\
            DAT & Semantic-FPN & 32.3 & 198.8 & 42.63 & 54.72 \\
            \cellcolor[RGB]{232,240,254}\textbf{AnchorFormer} & \cellcolor[RGB]{232,240,254}Semantic-FPN & \cellcolor[RGB]{232,240,254}21.6 & \cellcolor[RGB]{232,240,254}193.7 & \cellcolor[RGB]{232,240,254}\textbf{43.81} & \cellcolor[RGB]{232,240,254}\textbf{56.57} \\
            \midrule
            Swin & UperNet & 57.3 & 945.8 & 45.28 & 54.32 \\
            Focal & UperNet & 62.3 & 998.1 & 45.8 & 52.62 \\
            RMT & UperNet & 56.4 & 937.4 & 49.8 & - \\
            GraftViT & UperNet & 66.4 & 954.6 & 45.5 & 53.2 \\
            \cellcolor[RGB]{232,240,254}\textbf{AnchorFormer} & \cellcolor[RGB]{232,240,254}UperNet & \cellcolor[RGB]{232,240,254}49.6 & \cellcolor[RGB]{232,240,254}928.1 & \cellcolor[RGB]{232,240,254}\textbf{46.28} & \cellcolor[RGB]{232,240,254}\textbf{56.33} \\
            \bottomrule
    \end{tabular}}
    
    \label{semantic_result}
\end{table}

\subsection{Object Detection}
Meanwhile, to verify the efficiency of the downstream object task, we extend AnchorFormer on the COCO dataset and introduce some efficient detection models as the baselines. Specifically, we employ ESNet \cite{yu2111pp} as the backbone and replace the last stage with the proposed AnchorFormer. Meanwhile, the detector head and the training settings are followed by PicoDet. The comparative results are listed in Table \ref{Object_detect_result}. We can find that ours consistently obtains the best trade-off between accuracy and efficiency, i.e., $\uparrow \bm{30.0\%} \sim \uparrow \bm{81.3\%}$, $\uparrow \bm{6.4\%} \sim \uparrow \bm{35.3\%}$, and $\uparrow \bm{49.9\%} \sim \uparrow \bm{67.3\%}$ mAP improvements compared to YOLO, EfficientDet and FBNetV5, respectively, under comparable FLOPs. Our mAP is far higher than the detection based on MobileNet under comparable parameters. Especially for the baselines of YOLO and MobileNet, the proposed AnchorFormer-T can achieve the highest mAP with the smallest FLOPs. It is mainly because that the proposed model can differentiable learn the pivotal information distributed randomly among the image dataset. Thus, we can achieve the best trade-off performance and efficiency on the downstream object detection task. 

\begin{table}[t]
    \renewcommand\arraystretch{1.2}
    \centering
    \caption{Ablation study of the proposed AnchorFormer with different designed parts. The ViT means the vanilla ViT baseline. They are conducted on the ImageNet dataset for classification. The last column represents the Top-1 improvements of each model than DeiT-T baseline.}
    \vspace{-1mm}
    \scalebox{0.8}{
        \begin{tabular}{p{1.1cm}<{\centering}p{1.1cm}<{\centering}p{1.1cm}<{\centering}|cc|cc}
            \toprule
            \textbf{\begin{tabular}[c]{@{}c@{}}ViT \end{tabular}} 
            & \textbf{\begin{tabular}[c]{@{}c@{}}Anchor \end{tabular}} 
            & \textbf{\begin{tabular}[c]{@{}c@{}}Diff. \end{tabular}} 
            & \textbf{\begin{tabular}[c]{@{}c@{}}Params\\ (M)\end{tabular}} 
            & \textbf{\begin{tabular}[c]{@{}c@{}}FLOPs\\ (G)\end{tabular}} 
            & \textbf{\begin{tabular}[c]{@{}c@{}}Top-1\\ (\%)\end{tabular}} 
            & \textbf{\begin{tabular}[c]{@{}c@{}}Improv.\\ (\%)\end{tabular}} 
            \\
            \midrule
            \checkmark & & & 6.3 & 1.5 & 72.51 & 2.58 ({\color{green}$\downarrow$})\\
            \checkmark & \checkmark & & 4.7 & 0.9 & 77.48 & 4.00 ({\color{black}$\uparrow$})\\
            \checkmark & \checkmark & \checkmark & \cellcolor[RGB]{232,240,254}4.2 & \cellcolor[RGB]{232,240,254}0.7 & \cellcolor[RGB]{232,240,254}81.12 & \cellcolor[RGB]{232,240,254}\textbf{8.31} ({\color{black}$\uparrow$}) \\ \midrule
            \multicolumn{3}{c|}{DeiT-T} & 5.6 & 1.3 & 74.38 & 0.00 ({\color{black}$\uparrow$})\\
            \bottomrule
    \end{tabular}}
    
    \label{ablation_result}
\end{table}

\begin{table}[t]
    \renewcommand\arraystretch{1.2}
    \centering
    \caption{Ablation study of the proposed AnchorFormer with the different number of anchors. The experiments are conducted on DeiT-S and DeiT-B for ImageNet classification. The last column represents the Top-1 improvements of each model than DeiT-T and DeiT-S baselines, respectively.}
    \vspace{-1mm}
    \scalebox{0.8}{
        \begin{tabular}{p{0.7cm}<{\centering} p{0.7cm}<{\centering} p{0.7cm}<{\centering} p{0.7cm}<{\centering}|cc|cc}
            \toprule
            \multicolumn{4}{c|}{\textbf{Anchor Nums.}} 
            & \textbf{\multirow{2}{*}{\begin{tabular}[c]{@{}c@{}}Params\\ (M)\end{tabular}}} 
            & \multicolumn{1}{l|}{
                \textbf{\multirow{2}{*}{\begin{tabular}[c]{@{}c@{}}FLOPs\\ (G)\end{tabular}}}
            } 
            & \textbf{\multirow{2}{*}{\begin{tabular}[c]{@{}c@{}}Top-1\\ (\%)\end{tabular}}} 
            & \textbf{\multirow{2}{*}{\begin{tabular}[c]{@{}c@{}}Improv.\\ (\%)\end{tabular}}}
            \\
            \cline{1-4}
            \multicolumn{1}{c|}{\textbf{10}} & \multicolumn{1}{c|}{\textbf{30}} & \multicolumn{1}{c|}{\textbf{50}} & \multicolumn{1}{c|}{\textbf{100}} &                                                                
            & \multicolumn{1}{c|}{}                                                               
            &                                                                
            &                                                                \\
            \midrule
            \checkmark & & & & 3.9 & 0.6 & 77.25 & 3.72 ({\color{black}$\uparrow$})\\
            & \checkmark & & & \cellcolor[RGB]{232,240,254}4.2 & \cellcolor[RGB]{232,240,254}0.7 & \cellcolor[RGB]{232,240,254}81.12 & \cellcolor[RGB]{232,240,254}\textbf{8.31} ({\color{black}$\uparrow$})\\
            & & \checkmark & & 4.3 & 0.8 & 79.62 & 6.58 ({\color{black}$\uparrow$})\\
            & & & \checkmark & 4.7 & 1.0 & 75.83 & 1.91 ({\color{black}$\uparrow$})\\ 
            \hline
            \multicolumn{4}{c|}{DeiT-T} & 5.6 & 1.3 & 74.38 & 0.00 ({\color{black}$\uparrow$})\\
            \midrule
            \checkmark & & & & 15.3 & 2.4 & 81.73 & 0.18 ({\color{black}$\uparrow$})\\
            & \checkmark & & & \cellcolor[RGB]{232,240,254}18.6 & \cellcolor[RGB]{232,240,254}2.7 & \cellcolor[RGB]{232,240,254}84.69 & \cellcolor[RGB]{232,240,254}\textbf{3.67} ({\color{black}$\uparrow$})\\
            & & \checkmark & & 19.1 & 2.9 & 82.16 & 0.71 ({\color{black}$\uparrow$})\\ 
            & & & \checkmark & 19.4 & 3.2 & 80.69 & 1.10 ({\color{green}$\downarrow$})\\ 
            \hline
            \multicolumn{4}{c|}{DeiT-S} & 21.9 & 4.6 & 81.58 & 0.00 ({\color{black}$\uparrow$})\\
            \bottomrule
        \end{tabular}
    }
    
    \label{ablation_anchors}
\end{table}

\subsection{Semantic Segmentation}
Furthermore, we extend the proposed AnchorFormer on the semantic segmentation task and verify the performance and efficiency.  We employ the Semantic-FPN \cite{kirillov2019panoptic} and UperNet \cite{xiao2018unified} as the backbones on ADE210K. As shown in Table \ref{semantic_result}, the proposed model achieves better performance than the baselines in terms of accuracy and efficiency. Among them, AnchorFormer obtains $\downarrow\bm{14.4\%}$ FLOPs reduction and $\uparrow \bm{2.3\%}$ mIoU improvement than PVT. The proposed model obtains $\downarrow\bm{1.88\%}$ FLOPs reduction and $\uparrow \bm{2.2\%}$ mIoU improvement than Swin. 
This experiments proves that AnchorFormer can be generalized to various vision tasks and achieve excellent performance.

\subsection{Ablation Study}
To validate the effectiveness of anchor vision transformer (\textbf{Anchor}) and differentiable anchor vision transformer (\textbf{Diff.}), we conduct several ablation studies on image classification tasks. 
As shown in Table \ref{ablation_result}, the differentiable anchor vision transformer achieves better performance than the others in terms of accuracy and efficiency. 
{\color{black}
Compared with the basic anchor vision transformer, the differentiable model obtains $\downarrow\bm{22.2\%}$ FLOPs reduction and $\uparrow \bm{4.5\%}$ top-1 accuracy improvement. It suggests that the proposed model can accurately and differentiable learn the pivotal information among the input image dataset.
}
Besides, we also conduct ablation experiments to study the sensitivity of the different numbers of anchors. DeiT-T and DeiT-T are employed and the number of anchors is selected from $[10, 30, 50, 100]$. 
As shown in Table \ref{ablation_anchors}, more and less number of anchors may not achieve the highest performance on ImageNet classification. 
{\color{black}
Specifically, under comparable FLOPs, AnchorFormer with $30$ anchors obtains $\uparrow\bm{4.8\%}$ and $\uparrow\bm{3.5\%}$ than AnchorFormer with $10$ on DeiT-T and DeiT-S, respectively. It means that too less anchors may discard some features and drop the performance. AnchorFormer with $30$ anchors achieves FLOPs reduction and top-1 accuracy improvement simultaneously than AnchorFormer with $100$ anchors. It means that too many anchors also introduce redundant information to limit the performance.  Therefore, we can simply set the anchor number as a median like $30$.
}


\section{Conclusion}
In this paper, we propose an anchor-based efficient vision transformer to learn the pivotal information with the anchor tokens and accelerate the inference of ViTs. 
It mainly estimates the bipartite attention between the anchors and the tokens to reduce the complexity. 
Thus, we can differentiable learn global self-attention through the Markov process and the complexity will be reduced to $\mathcal{O}(mn)$, where $m$ is an anchor number and $m < n$. 
Moreover, extensive experiments can verify the performance and efficiency of our model. 
\textcolor{black}{
It is the main limitation that AnchorFormer seems to be incompatible with the causal attention, which is widely used in the modern LLMs. 
It seems to limit the extension of AnchorFormer to natural language processing. 
}
However, we believe that the proposed anchor-based strategy opens up a new perspective on efficiency improvement. 
In the future, we may apply the proposed AnchorFormer to ancient text recognition and analysis 
as well as image analysis and classification of artifacts \cite{Ioannis1} in archeological research \cite{Ioannis2} and STEM in Arts and Culture \cite{Ioannis3}, owing to its remarkable efficiency.
It is also promising to further improve the proposed method by using positive-incentive noise theory \cite{PN} since 
the anchor attention can be regarded as a noisy approximation of the vanilla attention. 
The positive-incentive noise \cite{PN} is the first mathematical framework to quantify the noise impact. 
The novel concept shows us how to systematically study the noise and the following series of works \cite{VPN,PiNDA,PiNI,WhyDropEdge,PiNGDA} shows us how to effectively apply the elegant framework to the popular deep learning models. 

\bibliographystyle{elsarticle-num.bst}
\bibliography{main}
\end{document}